\documentclass[letterpaper, 10 pt, conference]{ieeeconf}  

\usepackage{multicol}
\usepackage{amssymb}
\usepackage[bookmarks=true]{hyperref}
\usepackage{wrapfig}
\usepackage{amsmath}
\usepackage{color,soul}
\usepackage{xcolor}
\usepackage[normalem]{ulem}
\usepackage{graphicx}

\IEEEoverridecommandlockouts                              

\title{\LARGE \bf
Neural Field Movement Primitives  \\  for Joint Modelling of Scenes and Motions
}

\author{Ahmet Tekden$^{1}$
\and Marc Peter Deisenroth$^{2}$
\and Yasemin Bekiroglu$^{1,2}$
    \thanks{This work was supported by Chalmers AI Research Center (CHAIR) and Chalmers Gender Initiative for Excellence (Genie), and partially supported by the Wallenberg AI, Autonomous Systems and Software Program (WASP) funded by the Knut and Alice Wallenberg Foundation. 
          $^{1}$Electrical Engineering, Chalmers University of Technology, Sweden. $^{2}$Computer Science, University College London, U.K. Email: {\tt\small tekden@chalmers.se}}
}

\begin{document}

\maketitle
\thispagestyle{empty}
\pagestyle{empty}

\begin{abstract}

This paper presents a novel Learning from Demonstration (LfD) method that uses neural fields to learn new skills efficiently and accurately. It achieves this by utilizing a shared embedding to learn both scene and motion representations in a generative way. Our method smoothly maps each expert demonstration to a scene-motion embedding and learns to model them without requiring hand-crafted task parameters or large datasets. It achieves data efficiency by enforcing scene and motion generation to be smooth with respect to changes in the embedding space. At inference time, our method can retrieve scene-motion embeddings using test time optimization, and generate precise motion trajectories for novel scenes. The proposed method is versatile and can employ images, 3D shapes, and any other scene representations that can be modeled using neural fields. Additionally, it can generate both end-effector positions and joint angle-based trajectories. Our method is evaluated on tasks that require accurate motion trajectory generation, where the underlying task parametrization is based on object positions and geometric scene changes. Experimental results demonstrate that the proposed method outperforms the baseline approaches and generalizes to novel scenes. Furthermore, in real-world experiments, we show that our method can successfully model multi-valued trajectories, it is robust to the distractor objects introduced at inference time, and it can generate 6D motions.

\end{abstract}

\section{Introduction}

Learning from demonstration (LfD) methods enable robots to bootstrap the acquisition of new skills necessary to complete tasks by leveraging expert-provided demonstrations~\cite{lfd_survey}. These methods are useful in tasks where modeling the motion with respect to scene variations is difficult. When the task is complex, and the associated scene can only be described using highly nonlinear task parameters, LfD can facilitate the acquisition of new skills. It offers insights into how the skill is affected by changes in the task parameters and enables easier generalization to previously unseen scene conditions.

\begin{figure}
    \centering
    \includegraphics[width=\linewidth]{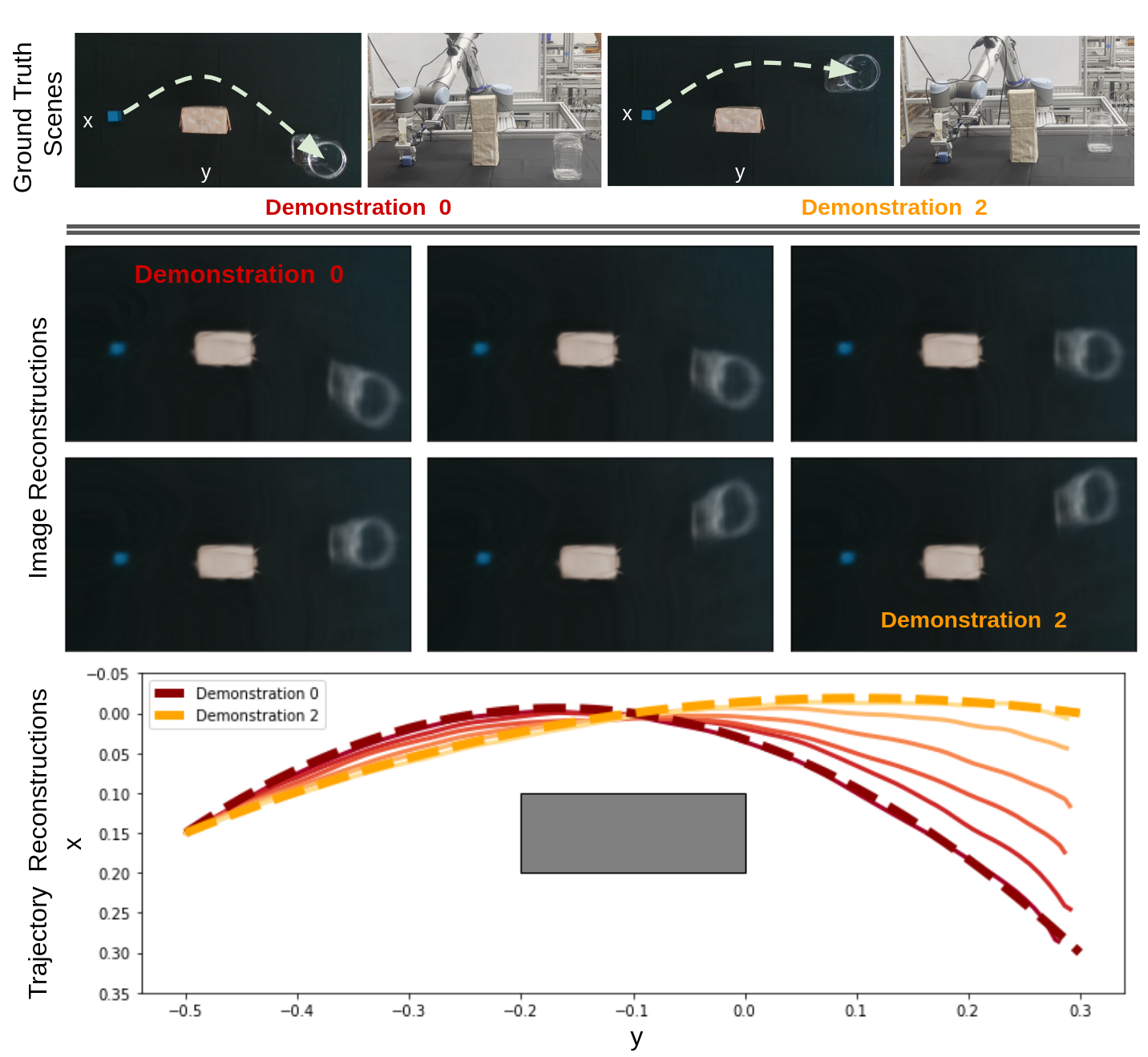}
    \caption{Examples illustrating the approach: Two expert demonstrations for two different plastic container locations are shown on top. Our network can smoothly and accurately interpolate between these demonstrations, generating the images and the trajectories corresponding to the unseen intermediate plastic container locations.}
    \label{fig:smooth_change}
\end{figure}

Movement primitives~\cite{dmp,promp,cnmp,vmp,acnmp,calinon2016tutorial} are often built upon LfD approaches that provide an efficient way to teach robots complex motor skills. In LfD, data efficiency corresponds to the number of expert demonstrations required to model the motion correctly. These methods~\cite{cnmp,acnmp,vmp,calinon2016tutorial} generally employ hand-crafted task parameters that well describe the scene to learn skills that generalize to novel scenes in a data-efficient way. Common examples of these task parameters are waypoints, obstacle locations, and sizes. Often, these task parameters have to be provided by the user, or they need to be estimated, which requires further engineering effort by experts that have access to clear knowledge of the tasks, and know the limitations of the underlying methods. To reduce the dependence on experts, some movement primitive architectures use raw images~\cite{cnmp,deep_dmp} to extract the factors that influence the skill. However, they usually require a thousand~\cite{cnmp}, or a few thousand~\cite{deep_dmp} demonstrations to achieve this. We propose an alternative approach: modeling the scenes and the motions using neural fields. Recently, it has been shown that neural fields~\cite{nerf_survey} can be utilized to represent 3D scenes~\cite{nerf}, shapes~\cite{deepsdf}, images~\cite{siren} using small amount of data. Inspired by this, we propose an LfD method that jointly models the scenes and the motions using neural fields. Our method\footnote{Project page: \url{https://fzaero.github.io/NFMP/}} learns to reconstruct the motion trajectories and the scenes given corresponding expert demonstrations of the task using neural fields. Later, it can generate accurate motion trajectories for novel scenes. An example generation of our method is shown in Figure~\ref{fig:smooth_change}. In this figure, our method interpolates between two expert demonstrations with different container locations and accurately generates the scene and the motion representations for unseen container locations. Unlike the previous works, our method achieves such trajectory generation without access to task parameter vectors at training or test time. Furthermore, it uses a similar number of expert demonstrations that the task parameter-based LfD methods use. 

Neural fields~\cite{nerf_survey} are powerful scene representation tools that can be employed to model images, surfaces, and 3D scenes. Using them allows for representing scenes in a format ideal for the task and provides a versatile way of modeling movement primitives. Neural fields can be equipped to handle modeling smooth changes in the scene by using deformation and template neural fields. Using deformation fields, our method maps smooth scene and motion changes to each other, enabling accurate motion trajectory generation for novel scenes. However, when there is a large change between samples, deformation fields fail to model changes in the scene correctly (\textit{i.e.,} rapid scene changes~\cite{nerfies}). This is because neural fields model complex signals through positional encodings~\cite{fourier_features}: Low-frequency positional encodings are good at modeling smooth changes, while high-frequency positional encodings are good at capturing fine details. Using low-frequency positional encodings results in overly smooth scene reconstruction. However, with high-frequency positional encodings, the model starts overfitting to high-frequency details in low-sample density settings and loses the generalization capacity. To prevent this from happening, we propose using low-frequency positional encodings for the deformation field while keeping the template field the same. This allows our network to capture the details necessary to represent the scene while still preserving the smoothness that low-frequency positional encodings provide. 

Our main contributions can be summarized as follows:
\begin{itemize}
    \item We propose a novel LfD method that simultaneously generates a scene and a motion representation by utilizing a shared embedding. Our method assigns an embedding vector to each expert demonstration and learns scene and motion neural fields smoothly parametrized with the embeddings. On novel scenes, by employing test time optimization to find an embedding that can correctly generate the scene, our approach can generalize to new scenarios and accurately predict the correct motion.
    \item Our method preserves the smooth changes in the scene by using low-frequency positional encodings for the deformation field. This allows our approach to be data-efficient while still capturing the necessary details to represent the scene. For all the experiments, the method generalizes to the novel scenes based on low number of demonstrations ($3^n$ demonstration for $n$ independent task parameters).
    \item We evaluate our approach both in simulation and real-world experiments. Simulation results show that our method outperforms the baseline movement primitive networks on the tasks where the underlying task parametrization is based on object positions and geometric scene changes. Our real-world experiments demonstrate that our method can handle multi-valued trajectories through implicit modeling of the motion. In addition, our method is robust to distractor objects introduced to the scene at the inference time, and it can model 6D pick and place tasks using a surface-based representation. 
\end{itemize}

\section{Related Work}
Learning from Demonstration (LfD) is a widely used approach in robot learning that leverages expert demonstrations to bootstrap the learning process~\cite{lfd_survey, schaal1996learning}. LfD~\cite{lfd_trend} can be employed to encode continuous joint or state trajectories, \textit{e.g.} movement primitives~\cite{dmp,promp,calinon2016tutorial, cnmp,vmp}; experience abstractions, \textit{e.g.} behavioral cloning~\cite{behavioral_cloning,ibc}, offline RL~\cite{offlinerl_tutorial}; and task waypoints, \textit{e.g.} affordance heatmaps~\cite{transporter_net}, keypoints~\cite{florence2018dense,NeRF-Supervision,NDF}. Our work is related to the first category, where the underlying robotic skills are encoded as low-level motion trajectories using neural fields. Methods in this category often use task parameters that describe the scene compactly to reduce the number of trajectories necessary to generalize to the variations in the scene~\cite{lfd_survey2}. However, this requires accurate extraction of task parameters from the scene both during the training and the test time. This can be challenging as it requires developing sophisticated techniques to estimate the task parameters for each different task. In practice, it is often handled by direct provision of the task parameters by a human operator. In contrast, our network directly works with any scene representation that can be generated through neural fields and is validated with RGB images and signed distance functions (SDF)~\cite{deepsdf}. Furthermore, by modeling the motions and the scenes through a shared embedding, our method learns movement primitives that smoothly generalize to changes in the task parameters. This allows our method to learn new skills with a similar number of demonstrations that previous task-parameter-based movement primitive methods use.

For representing scenes and motions, we use neural fields. Neural Fields~\cite{nerf_survey} are coordinate-based neural networks: They take continuous spatial or temporal coordinates as input to parametrize functions of such coordinates.  They allow for a representation of a wide variety of scene modalities, including images~\cite{siren}, implicit surfaces~\cite{deepsdf,occ-net}, 3D scenes~\cite{nerf}, point clouds~\cite{point-nerf}, tactile readings~\cite{virdo}, and even audio~\cite{siren}. Such representations have been used for SLAM~\cite{imap}, navigation~\cite{nerf-navigation}, control~\cite{implicit_vis_control}, and affordance estimation~\cite{mira}. Unlike these works, our method utilizes neural fields to generate scenes and their corresponding motion trajectories.    
\begin{figure*}
    \centering
    \includegraphics[width=\linewidth]{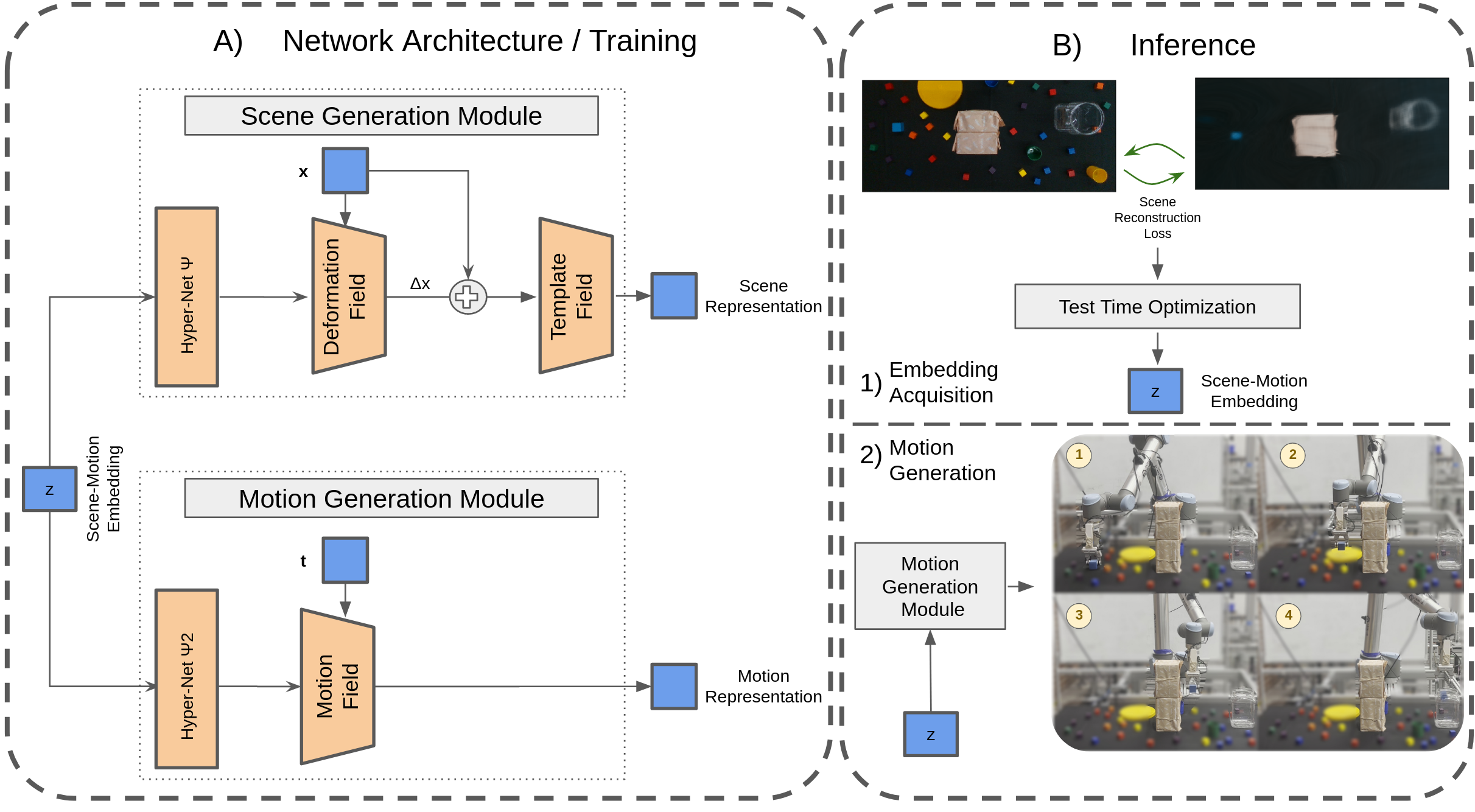}
    \caption{Our network maps scene-motion embeddings to scene representations, \textit{e.g.} RGB images or signed distance functions (SDF), and motion representations, such as joint angles or end-effector position of the robot (A). At inference time, through test-time optimization, scene-motion embeddings can be obtained through scene reconstruction loss (B-1) and can be used to reproduce trajectories for new scenes (B-2).}
    \label{fig:method}
\end{figure*}

Neural fields have also been used to map scenes to grasp locations~\cite{giga} or grasp trajectories~\cite{grasp_df1} using implicit functions. However, these networks employ large datasets that can only be generated via simulations. They learn the grasp trajectories from  millions of samples. In addition, designing a new environment for each unique task and bridging the simulation-to-reality gap are essential requirements for such methods. Instead, our method requires only a handful of demonstrations to model each new task by mapping scene changes to the motion. It can directly learn from real-world data, and it can still generate accurate motion trajectories. Other neural field-based approaches aim to model the scene changes smoothly through deformation fields~\cite{nerfies,dif-net}. However, these networks attain smoothness through denser data sampling. Collecting such densely sampled data in LfD settings is challenging as each scene variation requires an expert demonstration. Instead, we use low-frequency positional encodings to model smooth changes in a data-efficient way. 

\section{Method}

The overall framework is illustrated in Figure~\ref{fig:method}. Our network does not have traditional inputs and outputs. Instead, it assigns a scene-motion embedding for each expert demonstration. At training time, our method jointly learns to render the scene and reconstruct the motion for each provided demonstration, while optimizing their shared scene-motion embeddings (A). Our method can employ a wide range of scene and motion modalities that neural fields can model as long as their underlying representations are smooth. At inference time, for a given scene, the scene-motion embedding is first found through test-time optimization that minimizes the scene reconstruction loss (B-1). Later, this embedding is used to generate the corresponding motion trajectory (B-2). Section~\ref{sec:experiments} provides example tasks modeled using our method.

\subsection{Problem Formulation}

We are interested in LfD tasks parametrized with task parameter vector $p$ where the scene is represented with the function $f_p(x)$ with spatial coordinate $x\in \mathbb{R}^m$, and motion is represented with function $g_p(t)$ with normalized temporal coordinates $t\in[0,1]$. For these LfD tasks, $f_p(x)$ and $g_p(t)$ are smooth functions of $p$~\footnote{This is generally assumed in most LfD tasks.}.
\subsection{Neural Fields}
Neural fields are coordinate-based neural networks that take spatial or/and temporal coordinates and parametrize physical properties of the scene, objects, or motions. Scene and motion functions, $f_p(x)$ and $g_p(t)$, can be modeled as neural fields parameterized with task parameter vector $p$. Neural fields can model a wide range of scene and motion representations:
\paragraph*{Images} Images are common visual representations in robotics acquired using a wrist or a scene camera. Using coordinate-based MLPs, a mapping, $\mathbb{R}^2 \rightarrow\mathbb{R}^3$, between pixel locations and RGB values is learned to represent the scenes.
\paragraph*{Surfaces} 3D surfaces are represented with signed distance functions (SDFs) $\mathbb{R}^3\rightarrow \mathbb{R}^1$, which map 3D locations onto their corresponding signed distance value~\cite{deepsdf}. For a given point $x$, SDF yields the distance between the underlying surface and $x$, which is defined as $x<0$ if $x$ is inside the surface, and $x>0$ if $x$ is outside of the surface. Through the marching cubes algorithm~\cite{marching-cubes}, the 3D mesh of the scene is recovered using the signed distance values obtained from the SDF. 
\paragraph*{Motion Trajectory} The motion trajectory is the sequence of states of end effector poses or joint angles. Using normalized temporal coordinates, a mapping, $\mathbb{R}^1\rightarrow \mathbb{R}^n$, between time $t$ and $n$-dimensional joint/pose state $q$ is employed to model the motion. In the case of multi-valued trajectories, motion is represented implicitly with a mapping $\mathbb{R}^{n+1}\rightarrow \mathbb{R}^1$, where $(q,t)$ coordinates are mapped to state or cost values~\cite{ibc}.

In our framework, we adapt the neural field architecture proposed in  SIREN~\cite{siren}, and model the neural fields with multi-layer perceptron (MLP) parametrized using a Hypernet~\cite{hypernet}.

\subsection{Positional Encodings}
 
Using spatial or temporal coordinates directly in neural networks leads to blurry reconstructions~\cite{nerf}. To prevent this, positional encodings~\cite{fourier_features} are utilized. Specifically, for a coordinate $x\in\mathbb{R}^m$, we use the positional encoding map $
\Gamma(x) = [x,\Gamma_0(x),\Gamma_1(x),\ldots,\Gamma_{L-1}(x)]\in \mathbb{R}^{m+2mL}$, where $\Gamma_{n}(x)=[\cos(2^n \pi x),\sin(2^n \pi x)] \in \mathbb{R}^{2m}$ and $L$ is the number of frequency bands. Higher frequency bands are useful for representing high-frequency signals, such as fine details. In contrast, lower frequency bands are better at capturing low-frequency signals, such as the general silhouette of the scene.

\subsection{Scene-motion Embeddings}
Obtaining task parameters that describe the scene well is often challenging, even for the initially provided demonstrations. Instead, we assign a scene-motion embedding $z\in\mathbb{R}^k$ to each demonstration with an unknown task parameter vector $p$. We then represent the scene and the motion as smooth functions of embedding $z$. With this new formulation, tasks are equivalently represented with neural fields $F_z(x)$ and $G_z(t)$ parameterized by $z$ embedding.

\begin{equation}
    (F_z(x),G_z(t)) \equiv (f_p(x),g_p(t))
\end{equation}

In LfD, the number of available demonstrations is often small. This makes it challenging for an encoder network to learn a smooth mapping between the input scenes and the latent space for $z$. As a result, even slight changes in the scene may cause large changes in the motion at inference time, potentially resulting in incorrect motion predictions. Instead, we directly optimize the $z$ embeddings for each expert demonstration alongside the scene and motion neural fields. Test time optimization is utilized at inference time to predict the correct $z$ embedding for new scenes.

\subsection{Deformation and Template Fields}
In our framework, we assume the scenes are smooth functions of the task parameters, and the changes between the scenes can be expressed using deformations. With this assumption, we model the scene using a template neural field that models the canonical scene $T(x)$ and its deformation with a deformation field $D_z(x )=\Delta x$: 
\begin{equation}
    F_z(x) = T(x+D_z(x)).
\end{equation}
As the template field corresponds to the canonical scene and is not parametrized with $z$, it is modeled using a standalone MLP.

The deformation field causes an under-constrained optimization problem where different deformations could result in the same scene reconstruction. Increasing the sampling density of the scene with respect to the task parameters could alleviate this issue to some extent. However, as we would like to minimize the number of provided expert demonstrations, we rely on the following two approaches to address this problem:
\paragraph*{Coarse-to-fine Approximation}
In scene reconstruction with neural fields, using a small number of frequency bands admits low-resolution reconstruction, while a higher number of bands admit high-resolution reconstruction. In low-resolution reconstruction, it is easier to model larger changes in the scene; in high-resolution reconstruction, it is easier to model small scene changes. To balance this tradeoff, we use a coarse-to-fine approximation where the network first uses a low number of frequency bands, and the higher frequency bands are gradually introduced. This is achieved by applying a smoothing mask to the positional encodings. We adopt the coarse-to-fine approximation method proposed in the BARF~\cite{barf}.
\paragraph*{Deformation Fields with a Lower number of frequency bands} 
In our work, we model the template field using a high number of frequency bands and the deformation field using a low number of frequency bands. This allows our network to capture the details necessary to represent the scene while preserving the smoothness provided by the low-frequency positional encodings.
\subsection{Test Time Optimization}
We estimate the $z$ embedding through test time optimization. Given the scene image or the signed distance values of the modelled surface, we can find the embedding $z$ that minimizes the scene reconstruction loss  

\begin{equation}
    \text{argmin}_z \; L_{\text{scene}}(F_z,X)= \sum_{(x_j,s_j)\in X} \lVert F_z(x_j)-s_j\rVert_2^2 
\end{equation}

where $s_j$ corresponds to the ground truth scene signal at position $x_j$ for scene samples $X$. Ideally, the neural networks should establish a one-to-one mapping between the embeddings and the scene-motion functions. However, the network might learn embeddings that correspond to the same scene but with different motions. To mitigate this, we constrain the search space to the convex hull of embeddings utilized in the training process. If there are $n$ demonstrations in the training set, the optimization problem becomes

\begin{equation}
    \text{argmin}_\alpha \; L_{\text{scene}}(F_z,X) \text{ where }z = \sum_{i=0}^{n-1} \alpha_i z_i
\end{equation}

where $\alpha = [\alpha_0,...,\alpha_{n-1}]$ and $\sum_{i=0}^{i=n-1}\alpha_i = 1$; $\alpha_i \geq 0$. $z_i$ corresponds to the learned embedding of the expert demonstration $i$. With this equation, $z$ is not directly optimized but estimated as the weighted sum of demonstration embeddings.

\section{Implementation Details}
For all neural fields, a 3-layer Relu MLP with 128 hidden layer sizes and for Hypernets, a 2-layer Relu MLP with 64 hidden layer sizes are used. For template and motion fields, positional encodings with 8 frequency bands and for the deformation field, positional encoding with 2 frequency bands are employed. All scene-motion embeddings used in our experiments have a vector size of 128. All experiments are conducted with a UR10 robot equipped with an RG2 gripper and D435i wrist camera. Pybullet~\cite{pybullet} simulator is used for simulation experiments.

\begin{figure}
    \centering
    \includegraphics[width=\linewidth]{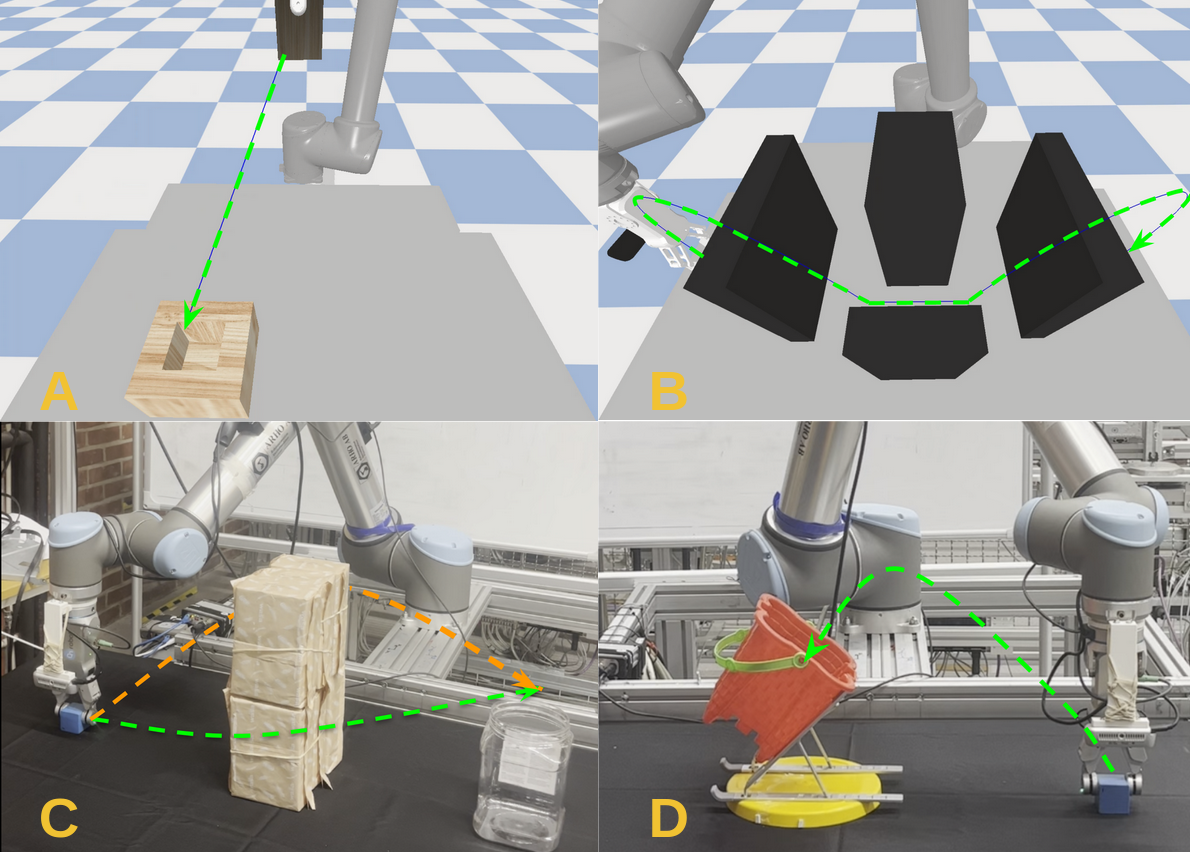}
    \caption{Tasks and the trajectory visualization of the expert demonstrations. Experiments for tasks A and B are conducted in simulation while others are conducted in real-world. For task C, each scene has two alternative motions shown with the color orange and light green.}
    \label{fig:tasks}
\end{figure}

We initialize the $z$ embedding for each demonstration as zero vectors. We use the learning rate $10^{-3}$ for network parameters, and $10^{-4}$ for embeddings during training. We use L2 loss between the predicted and the ground truth signal values for motion and scene reconstruction, and the square of the Euclidean norm of embeddings as regularization loss. To prevent learning unnecessary deformation mappings between background pixels, we utilize the square of the Euclidean norm of deformations. In summary, the loss function during training is formulated as:
\begin{equation*}
    w_1 L_{\text{motion}}+w_2 L_{\text{scene}}+w_3 L_{\text{deform\_reg}}+w_4 L_{\text{embedding\_reg}}
\end{equation*}
where $w$ are weights for each loss term. Values for $w_1$ and $w_2$ are chosen such that they balance each other during training. For $w_3$ and $w_4$, we choose the highest values that do not hinder the performance of the scene and motion reconstruction. All network parameters are fixed at the test time, and we optimize the $\alpha$ vector with a learning rate $0.05$ for each test demonstration only from the scene reconstruction loss $L_{\text{scene}}$. Adam~\cite{adam} optimizer is used both for training and test time optimization.

\begin{figure}
    \centering
    \includegraphics[width=\linewidth]{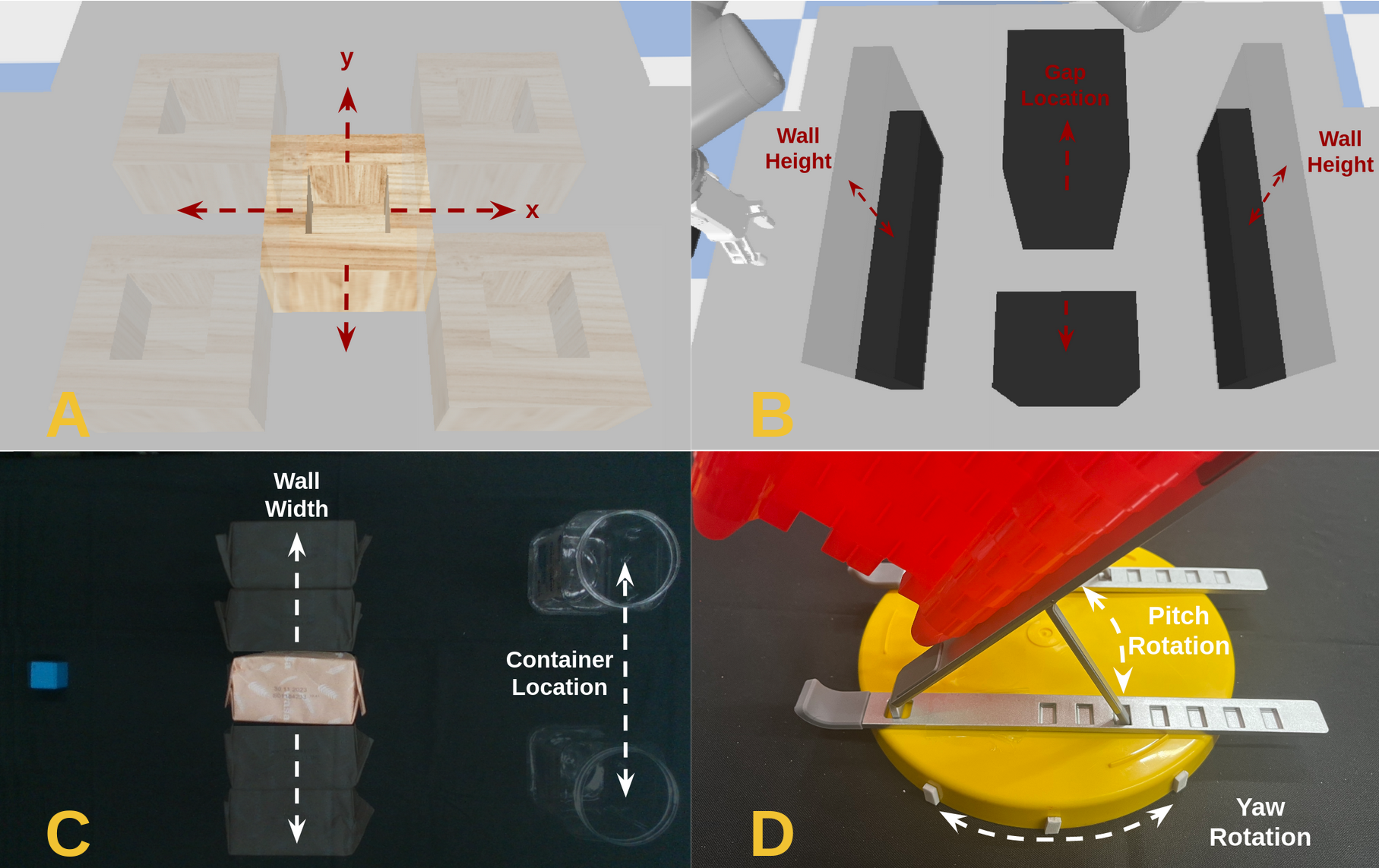}
    \caption{Illustration of how tasks are parameterized. Each arrow shows the extent and the boundary of each task parameter. Expert demonstrations for each task are collected from the boundary and the center locations of the task parameters. The generalization is tested on the interpolation region of these task parameters.}
    \label{fig:task_parameters}
\end{figure}

\section{Experiments}\label{sec:experiments}
To demonstrate the capabilities of our network, we design two simulation and two real-world robotic experiments. Visualizations of the experiment setups along with the demonstrated trajectories are shown in Figure~\ref{fig:tasks} and their details can be found in Table~\ref{table:tasks}. Furthermore,  Figure~\ref{fig:task_parameters} shows the task parameterization. These experiments demonstrate that our framework works on RGB and SDF-based scene representations and position- and joint-based motion representations. 

For a baseline comparison, we use two versions of a recent neural network-based motion primitive architecture, CNMP~\cite{cnmp}. These baselines are task parameter CNMP (TP-CNMP), which has direct access to task parameters, and C-CNMP, which processes the image of the scene via a CNN encoder to predict the motion trajectory. The network proposed in this paper outperforms the baseline architecture on generalization to shape and position in simulation experiments. In addition, as supported by the real robot experiments, the network can model multi-valued trajectories using implicit motion representation and enables 6D pick-and-place through 3D shape-based generalization.  

\begin{table}[!t]
\caption{Task Details}
\label{table:tasks}
\centering
\begin{tabular}{c||c|c}
Task Name  & Scene Modality & Motion Modality   \\
\hline\hline
Peg in hole    & RGB Image & Joint Angles \\ 
Wall Avoidance 1 & RGB Image & Joint Angles \\
Wall Avoidance 2 & RGB Image & Multi-valued Position \\
6D Pick and Place & Object Surface & Joint Angles \\
\end{tabular}
\end{table}
\subsection{Shape and Position Based Generalization}

To test our model's generalization capacity, we design two simulation experiments: based on position, and shape generalization. For both experiments, a scene camera is placed on the opposite side of the robot so that it can well capture the possible changes in the scene. 

For the position-based generalization, we set up a peg-in-hole experiment. This experiment contains a box with a peg slot put on a table at a given $(x,y)$ location. The task for the robot is to insert the peg into the slot on the box while generalizing to its location. A trajectory for the task is shown in Figure~\ref{fig:tasks}A. The task parameters for this task are the $x$ and $y$ location of the box. For the shape-based generalization, we set up a wall avoidance experiment with three walls. How this task is parameterized, and the trajectory that robot follows with its end effector is shown in Figure~\ref{fig:task_parameters}B, and Figure~\ref{fig:tasks}B. This task can be expressed most compactly with three task parameters: the first wall's height, the gap location on the second wall, and the third wall's height. 

\begin{figure}[!t]
    \centering
    \includegraphics[width=\linewidth]{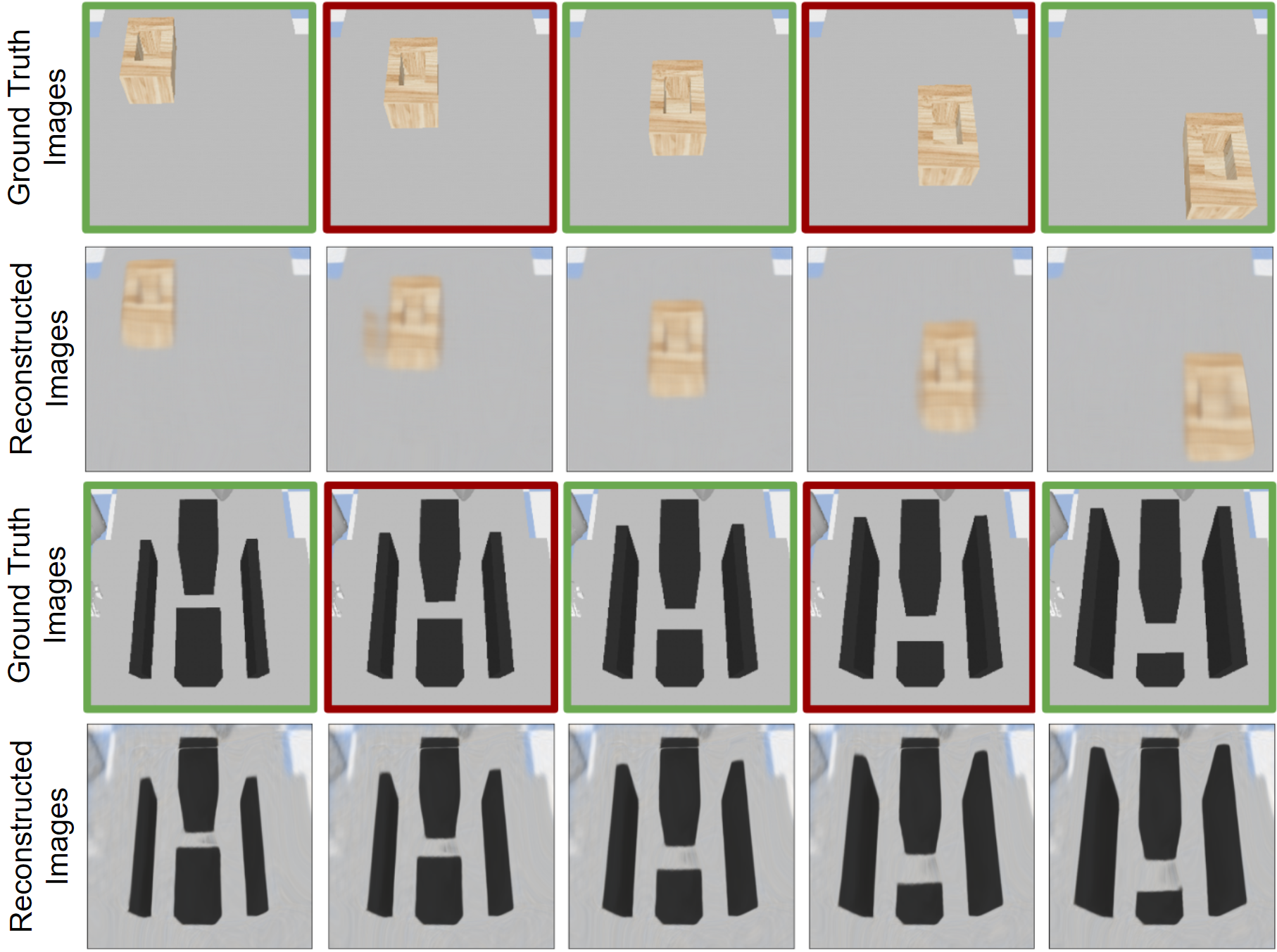}
    \caption{Comparison between reconstructed images and ground-truth images. Images with a red border are not seen during the training. However, the scene neural field can smoothly interpolate between the images seen in training and can reconstruct the unseen ones.}
    \label{fig:interpolations}
\end{figure}

For both experiments, joint angle-based motion representations, and RGB image-based scene representations are used. For generating the training datasets, expert demonstrations and the corresponding scene images are collected from the boundary and the center locations of the task parameters. These datasets are named 3x3 and 3x3x3 for peg-in-hole and wall-avoidance experiments, respectively, and contain 9 and 27 expert demonstrations. For evaluating our network, test datasets using mesh grids of the task parameters with grid sizes 4 and 5 are collected. These datasets contain data from the interpolation region of training demonstrations, and they aim to show how much our network can generalize to novel scenes. These datasets are named based on their grid configurations: 4x4, and 5x5 with 16 and 25 trajectories, respectively, for the peg-in-hole experiment, and 4x4x4 and 5x5x5 with 64 and 125 trajectories, respectively, for the wall-avoidance experiment.

Our comparison with the baseline architecture can be found in Table~\ref{generalization_results}. Our network outperforms both baselines in all cases. Task-parameter-based CNMP performs better than CNN-based one. However, even when the CNMP uses the task parameters directly, our method outperforms it, especially for the wall experiment, which involves more complex motion generation. This is because positional encoding not only improves the scene generation but also enhances the motion generation (\textit{i.e.,} without positional encoding, neural network generates overly-smooth trajectories). To illustrate our network's smooth interpolation capability, we further show our model's scene generation results from 5x5 and 5x5x5 datasets in Figure~\ref{fig:interpolations}. In the first and third rows, data points seen in training are shown with a green border, while unseen ones are shown with a red border. In the second and fourth rows, it can be seen that our network can successfully model shape and position changes between samples correctly. 

\begin{table}
\caption{Generalization Results (Average Joint Angle Error in $deg$)} 
\label{generalization_results}
\centering
\begin{tabular}{c||c|c|c}
\bfseries Test Case  & \bfseries Our Method & \bfseries TP-CNMP & \bfseries C-CNMP  \\
\hline\hline
3x3 & $\mathbf{0.94} $ & $1.01 $ & $1.24 $\\ 
4x4 & $\mathbf{2.40} $ & $2.55 $ & $3.03 $\\
5x5 & $\mathbf{2.28} $ & $2.60 $ & $3.04 $\\
\hline\hline
3x3x3 & $\mathbf{1.36} $ & $3.87 $ & $2.28 $\\ 
4x4x4 & $\mathbf{2.13} $ & $3.75 $ & $3.90 $\\
5x5x5 & $\mathbf{2.27} $ & $3.76 $ & $4.13 $\\
\end{tabular}
\end{table}

\subsection{Multi-valued Trajectory Generation}
This experiment shows that our method can model multi-valued trajectories by using implicit motion representations. The experiment includes several objects and obstacles: a blue cube on the left side of the table, a wall made of boxes in the middle, and a plastic container on the other side of the table. The task for the robot is to pick up the blue cube and drop it into the plastic container while avoiding the wall. However, this task can be achieved in two different ways, as illustrated in Figure~\ref{fig:tasks}C. The task is parameterized with the number of boxes in the wall, and the y location of the plastic container. The task is modeled using RGB images and position trajectories. Training data is collected using walls with one, three, and five boxes and a plastic container with three different locations, where $y$=\{$-0.15$, $0$, $0.15$\}. For each scene, one trajectory that goes from the left and one trajectory that goes from the right are recorded. In total, 18 demonstrations are collected.

\begin{figure*}
    \centering
    \includegraphics[width=\linewidth]{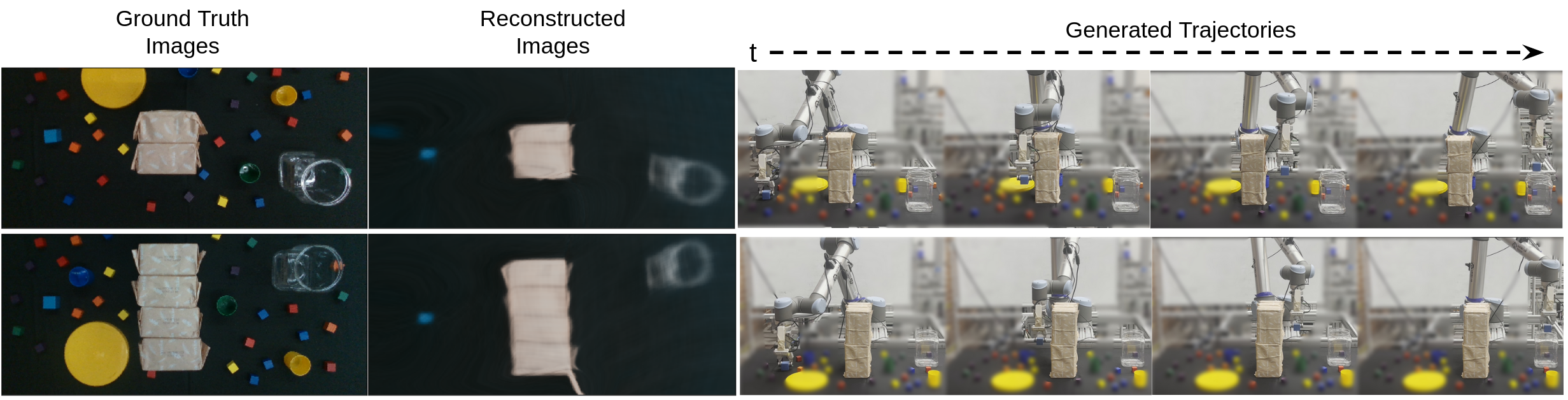}
    \caption{Illustration on the robustness of our method against distractor objects. Our method successfully ignores the distractor objects in its image reconstructions and can generate accurate motion trajectories for unseen environments.}
    \label{fig:exp3_distractors}
\end{figure*}

\begin{figure*}
        \centering
    \includegraphics[width=0.96\linewidth]{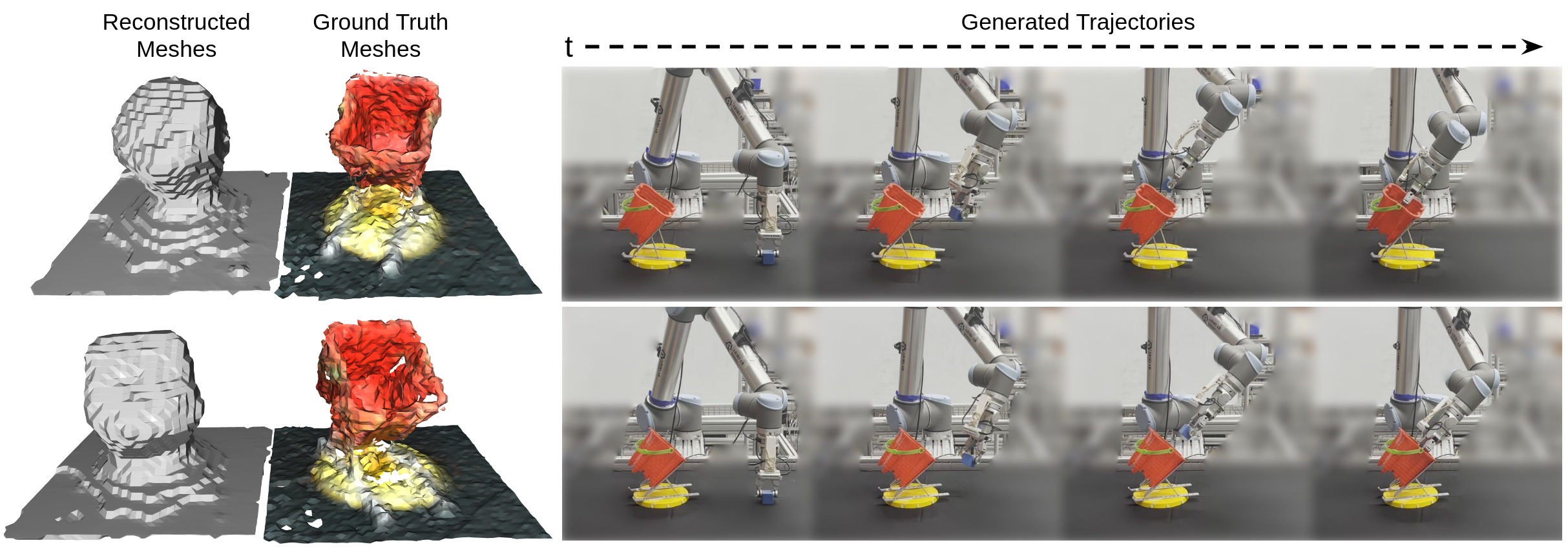}
    \caption{Illustration on how our method can utilize surface-based representations. While our method fails to model the opening of the box in its shape reconstruction due to the quality of the modeled mesh, it still manages to model the rotation of the box correctly. This allowed our method to generalize to novel scenes, and generate accurate motion trajectories for them.}
    \label{fig:exp4_reconstructions}
\end{figure*}

Since there are two alternative trajectories for each scene, we model the motion implicitly. This is achieved as follows: for a given 3D location in the scene, if it is on one of the ground truth trajectories, the cost value corresponds to 0; otherwise, its cost value is equal to its distance to the ground truth trajectories (for the distance calculation, the time coordinate is also included). However, implicit trajectory representations lack the same level of smoothness as explicit representations. Therefore, we similarly adopt deformation fields for representing the implicit motion. 

Since implicit motion only provides the cost function, the optimal trajectory is acquired by optimizing the following term:
\begin{equation*}
    \text{argmin}_{q}\int_{t=0}^1 G_z(q_t,t).
\end{equation*}
While previous movement primitive methods~\cite{promp,cnmp} require conditioning to handle multi-valued trajectories, our method can directly generate optimal trajectories without such conditioning. Nevertheless, if required, soft conditioning of the trajectory can be achieved by initializing $q$ close to the desired trajectory mode.       

We evaluate our method with $12$ unseen environments, with walls made of two and four boxes, and novel plastic container locations. To further test our method's robustness, distractor objects are added to the scene. These distractor objects and how our method ignores them in its scene reconstructions and successfully generates motion trajectories are shown in Figure~\ref{fig:exp3_distractors}.  For these $12$ unseen environments, our method accurately generates correct trajectories and acquires 100 \% accuracy. Furthermore, Figure~\ref{fig:smooth_change} illustrates the smooth interpolation of scenes and trajectories between demonstrations $0$ and $2$. Overall, our method can accurately model this task. 

\subsection{Shape-based motion generation}
In this experiment, we show that our model can exploit surface-based 3D shape representations, which do not have a structured rasterized format, and enable performing a 6D pick-and-place task.

For this experiment, the task is to pick up the blue cube and place it in the box as shown in Figure~\ref{fig:tasks}D. This box is placed with yaw and pitch rotations shown in Figure~\ref{fig:task_parameters}D. As this box is angled, the robot needs to adapt its trajectory according to the rotation of the box. Moreover, the opening of the box is narrow, and the task requires around $2$\,cm precision to prevent the robot from colliding with the sides of the box. 

The task is modeled using signed distance values and joint angle trajectories. For acquiring the ground truth scene representation, a set of points with their corresponding signed distance values are sampled from the mesh of the box estimated by recording the scene from 8 pre-defined camera poses. For this process, open3D~\cite{open3d} and pointcloud-utils~\cite{point-cloud-utils} libraries are employed. The training dataset is collected on scenes with yaw rotations angles of $-30^o$, $0^o$, and $30^o$ and pitch rotations corresponding to setting the stand on the first, third, and fifth gaps. 

Similarly to our previous experiment, we evaluate our method's performance on $12$ unseen environments with novel yaw and pitch rotations. Our method generates accurate trajectories for each scene and manages to acquire 100 \% accuracy. However, since the estimated mesh is not watertight, meshes reconstructed using our method cannot model the opening of the box. Yet, since it can accurately model the rotation of the box, it still generates correct trajectories. Examples of estimated and reconstructed meshes, and their corresponding motion trajectories are shown in Figure~\ref{fig:exp4_reconstructions}.    

\section{Conclusion}
In this work, we present a novel LfD method that can simultaneously model the scene and the motion for a task. Our experiments show that our method can generalize to novel scenes by accurately generating their motion trajectories and reconstructing their underlying scene representations. In addition, the proposed approach outperforms the baseline methods, can use multiple different scene modalities, such as RGB images and SDF, can model multi-valued trajectories, and is robust to disturbances. In future work, we plan to increase our method's data efficiency further and enable effect-based conditioning for trajectory generation. Furthermore, we plan to validate our method in a wider variety of task settings, including dexterous manipulation.
\bibliographystyle{IEEEtran}
\bibliography{IEEEabrv,references}
\end{document}